\documentclass{ecai} 

\usepackage{latexsym}
\usepackage{amssymb}
\usepackage{amsmath}
\usepackage{amsthm}
\usepackage{booktabs}
\usepackage{enumitem}
\usepackage{graphicx}
\usepackage{color}
\usepackage{multirow}
\usepackage{caption}
\usepackage{subcaption}
\usepackage{enumitem}
\setlist{nosep}

\newcommand{\BibTeX}{B\kern-.05em{\sc i\kern-.025em b}\kern-.08em\TeX}

\begin{document}


\begin{frontmatter}


\paperid{123} 


\title{Label Attention Network for Temporal Sets Prediction: You Were Looking at a Wrong Self-Attention}


\author[A, B]{\fnms{Elizaveta}~\snm{Kovtun}\thanks{Corresponding Author. Email: elizaveta.kovtun@skoltech.ru}}
\author[A, C]{\fnms{Galina}~\snm{Boeva}}
\author[A, C]{\fnms{Andrey}~\snm{Shulga}} 
\author[A, D]{\fnms{Alexey}~\snm{Zaytsev}} 

\address[A]{Skolkovo Institute of Science and Technology}
\address[B]{Sber AI Lab}
\address[C]{Moscow Institute of Physics and Technology}
\address[D]{Risk Management, Sber}

\begin{abstract}
Most user-related data can be represented as a sequence of events associated with a timestamp and a collection of categorical labels. For example, the purchased basket of goods and the time of buying fully characterize the event of the store visit. Anticipation of the label set for the future event called the problem of temporal sets prediction, holds significant value, especially in such high-stakes industries as finance and e-commerce. A fundamental challenge of this task is the joint consideration of the temporal nature of events and label relations within sets. The existing models fail to capture complex time and label dependencies due to ineffective representation of historical information initially. We aim to address this shortcoming by presenting the framework with a specific way to aggregate the observed information into time- and set structure-aware views prior to transferring it into main architecture blocks. Our strong emphasis on input arrangement facilitates the subsequent efficient learning of label interactions. The proposed model is called Label-Attention NETwork, or LANET. We conducted experiments on four different datasets and made a comparison with four established models, including SOTA, in this area. The experimental results suggest that LANET provides significantly better quality than any other model, achieving an improvement up to $65 \%$ in terms of weighted F1 metric compared to the closest competitor. Moreover, we contemplate causal relationships between labels in our work, as well as a thorough study of LANET components' influence on performance. We provide an implementation of LANET to encourage its wider usage.

\end{abstract}


\end{frontmatter}


\section{Introduction}
 
Numerous domains, such as banking, the grocery industry, etc., treat data as event sequences. For example, in the financial industry, much attention is paid to the history of human banking transactions ~\cite{babaev2022coles,bazarova2024universal} or the history of purchases in e-commerce~\cite{sun2019bert4rec}. A common problem for event sequences is the prediction of the label for the next event based on the available history \cite{kang2018self, sun2019bert4rec}.

\begin{figure}[t] 
\includegraphics[width=\linewidth]{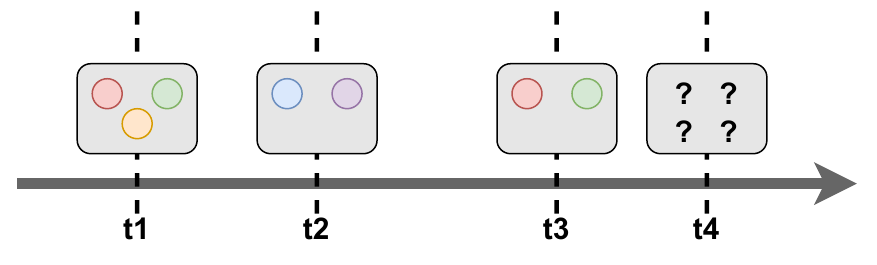} 
\caption{Visual representation of \textbf{temporal sets prediction problem}. The sequence of events that are characterized by timestamps $t_1, t_2, t_3$ and an arbitrary number of labels denoted with colored circles. Our goal is to \textcolor{blue}{predict label set for the next event} based on the previous sets.}
\label{fig:general_scheme}
\vspace{10pt}
\end{figure}

A natural extension of event sequences is temporal set data. For them, we observe a series of timestamped sets, where each set is composed of an arbitrary number of labels, see Figure~\ref{fig:general_scheme}. A primary goal is to predict the next set of labels. The difficulty lies in simultaneously accounting for the temporal sequential behavior of events and labels' interdependencies within sets. Understanding the composition of an expected event allows one to plan more accurately and, as a result, better manage resources.


Generally, the multi-label classification is a more natural setting than a binary or multiclass classification since everything that surrounds us in the real world is usually described with multiple labels~\cite{liu2021emerging}. 
There are numerous approaches to deal with the multi-label classification in computer vision~\cite{durand2019learning}, natural language processing~\cite{xiao2019label}, or classic tabular data domains~\cite{tarekegn2021review}. Temporal sets prediction can be viewed as multi-label classification problem for consecutive events.

The interaction between an object's states at different timestamps assists in solving tasks with sequential data~\cite{hartvigsen2020recurrent}. Therefore, expressive and powerful models should be able to learn such interactions. Several neural network architectures, such as transformers~\cite{vaswani2017attention} or recurrent neural networks~\cite{cho2014learning}, can do this. For example, a transformer directly defines an attention mechanism that measures how different timestamps in a sequence are connected. However, the applications of modern deep learning methods are limited~\cite{zhang2020multi}, and they primarily focus on predicting labels for a sequence in general.  

We refer to the graph of connections between states of an object at different timestamps as \textit{a timestamp interaction graph}. Another connection worth exploring is the connection between different labels and a need to consider the correlation between them~\cite{hang2021collaborative}. This capability is absent in the majority of models. 
We name the graph of connections between different labels \textit{a label interaction graph}.
 
In our research, we take into account both interaction between labels and timestamped events~\cite{li2020time,ying2018sequential}. For temporal sets prediction, simultaneous consideration of both \textit{timestamp interaction graph} and \textit{label interaction graph} is crucial. Typically, articles explore only one side of the dependence that can be explainable by domain bias. In sequential recommendation systems, there is a focus on connections between labels~\cite{quadrana2018sequence} with the incorporation of convolutional neural networks~\cite{tang2018personalized} as well as the attention mechanism~\cite{zhang2019next}.
Direct models for event sequences~\cite{hawkes1971spectra} prefer the identification of interactions between timestamps~\cite{zhuzhel2023continuoustime}, considering \textit{a timestamp interaction graph}.
 
\begin{table}[t!]
\centering
\caption{Mean rank for different metrics averaged over $4$ considered datasets. We want to minimize rank, as the best method has a rank of $1$. F1 and ROC AUC metric without specification refers to Weighted F1 and Weighted ROC AUC. We denote Hamming Loss as H Loss. The best values are in \textbf{bold}, and the second best values are \underline{underlined}.}
\vspace{2pt}
\begin{tabular}{p{1.2cm}p{1.1cm}p{1.1cm}p{0.6cm}p{1.3cm}p{1.0cm}}
\hline
Model & Micro F1 & Macro F1 & F1 & ROC-AUC & H Loss\\ \hline
SFCNTSP & 4.75 & 4.75 & 4.75 & 4.00 & 4.0\\ 
DNNTSP & 3.75 & 3.75 & 3.75 & 3.75 & 3.0\\  
GPTopFreq & 3.00 & \underline{2.75} & 3.00 & 4.25 & 3.5\\ 
TCMBN & \underline{2.50} & \underline{2.75} & \underline{2.50} & \underline{2.00} & \textbf{1.75} \\ 
LANET & \textbf{1.00} & \textbf{1.00} & \textbf{1.00} & \textbf{1.00} & \underline{2.25} \\ \hline
\end{tabular} 
\label{table:rank_results}
\end{table}
Our LANET aims at conjugate recovery of \textit{label interaction graphs} and \textit{a timestamp interaction graph}, as we believe it is a key moment for modeling temporal sets.
The algorithm aggregates past information in specially constructed representations. This aggregation phase is a distinctive feature of LANET that enables it to stand out among others. The built views serve as input to a transformer encoder. The encoder updates embeddings via self-attention, promoting learning of time and label interactions. Finally, we predict a vector of confidence scores for the next-event set based on the model output that encompasses deep knowledge of label relationships. Moreover, we can process long sequences this way, as now the attention evaluation is quadratic in the number of labels, not the sequence length.

\begin{figure*}[h!]
    \centering
    \includegraphics[scale=0.65]{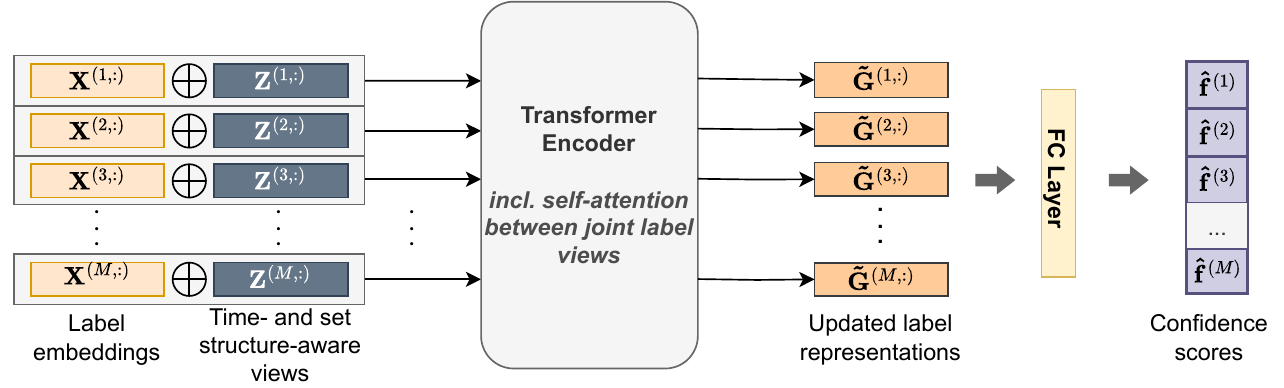}
    \vspace{5pt}
    \caption{LANET architecture for temporal sets prediction. The key part is to aggregate historical information into representative views that will be transferred into the Transformer encoder block. The output of the model is a vector of confidence scores, whose components are associated with the prospect of a corresponding label to be a member of the next-event set.}
    \label{fig:transf}
    \vspace{10pt}
\end{figure*}

\paragraph{\textbf{Contributions.}}
We propose a transformer-based architecture, called LANET, to effectively deal with temporal sets predictions. Our main contributions are the following:
\begin{itemize}
    \item We introduce LANET architecture for predicting a label set for the next event, taking the information from previous events. The architecture's peculiarity is based on the specific preparation of the historical information before transferring it into the block with self-attention. The scheme of our approach is presented in Figure~\ref{fig:transf}.
    \item We conduct a comprehensive comparison of LANET with the well-proven existing models for temporal sets prediction. All experiments indicate that LANET, due to its sophisticated input arrangement, outperforms all considered models by a large margin. See Table~\ref{table:rank_results} for a high-level comparison of different approaches.
    \item We study the influence of LANET components on its performance. The results suggest that LANET concentrates more on label linkages, while temporal information is in second place by importance.
\end{itemize}

\section{Related Work}
\label{sec:related}

Temporal sets prediction resembles a multi-label problem. The multi-label classification problem statement emerges in many diverse domains, e.g., text categorization or image tagging, all of which entail their peculiarities and challenges. The review~\cite{zhang2013review} explores foundations in multi-label learning, discussing the well-established methods as well as the most recent approaches. Emerging trends are covered in a review~\cite{liu2021emerging}.

We have identified several of the most relevant parts when studying this area. These sections describe significant features and approaches in the most detailed way. Firstly, we examine loss functions tailored for the multi-label setting and some methods for composing label set prediction. 
Secondly, we overview the usage of RNNs in the multi-label classification task. Thirdly, we review how to capture label dependencies. Then, we discuss an association with a sequential recommendation problem and next basket recommendation.


\paragraph{\textbf{Loss functions and ways for label set composition in multi-label problem.}}
The paper~\cite{menon2019multilabel} studies the theoretical background for main approaches to reducing a multi-label classification problem to a series of binary or multi-class problems. In particular, they show that considered reductions are implicitly optimized for either Precision@k or Recall@k. The choice of the correct reduction should be based on the ultimate performance measure of interest. In \cite{Li_2017_CVPR}, the authors propose an improved loss function for pairwise ranking in a multi-label image classification task that is easier to optimize. Also, they discuss an approach based on the estimating of the optimal confidence thresholds for the label decision part of the model that determines which labels to include in the final prediction. The task of multi-label text classification is the topic of \cite{du2019ml}.
The authors construct an end-to-end deep learning framework called ML-Net. ML-Net consists of a label prediction network and a label count prediction network. In order to get the final set of labels, confidence scores generated from the label prediction network are ranked, and then the top $K_{top}$ labels are predicted. A separate label count network predicts $K_{top}$. 

\paragraph{\textbf{Neural networks for multi-label classification.}}
In~\cite{yazici2020orderless}, the authors use the RNN model to solve a multi-label classification problem. The authors propose to dynamically order the ground truth labels based on the model predictions, which contributes to faster training and alleviates the effect of duplicate generation. In turn, \cite{tsai2020order} considers the transforming of a multi-label classification problem into a sequence prediction problem with an RNN decoder. They propose a new learning algorithm for RNN-based decoders that does not rely on a predefined label order. Consequently, the model explores diverse label combinations, alleviating the exposure bias. 
The work~\cite{shou2023concurrent} examines the same problem statement of multi-label classification in an event stream as we do. The authors' model targets capturing temporal and probabilistic dependencies between concurrent event types by encoding historical information with a transformer and then leveraging a conditional mixture of Bernoulli experts. This article~\cite{yu2023continuous} discusses the formulation of the task of predicting time sets for users; it offers a continuous learning system that allows you to explicitly capture changing user preferences by maintaining a memory bank that could store the states of all users and items. In this paradigm, the authors construct a non-decreasing universal sequence containing all user-defined interactions, then chronologically learn from each interaction. To research the cross-relation between products in the basket, a ConvTSP~\cite{zhang2023conv} was proposed that combines dynamic user interests and statistical interests into a single vector representation for a user.

\paragraph{\textbf{Approaches to leveraging label dependencies.}}
The authors of \cite{Lanchantin_2021_CVPR} construct a model called C-Tran for a multi-label image classification task that leverages Transformer architecture that encourages capturing the dependencies among image features and target labels. The key idea is to train the model with label masking. The authors in~\cite{yeh2017learning} propose DNN architecture for solving the multi-label classification task, which incorporates the construction of label embeddings with feature and label interdependency awareness. A label-correlation sensitive loss improves the efficiency of the constructed model. Another popular way to consider label relationships is to use Graph Neural Networks as a part of the pipeline. Namely, \cite{pal2020multi} captures the correlation between the labels in the task of Multi-Label Text Classification by adopting a Graph Attention Network (GAT). They predict the final set of labels combining feature vectors from BiLSTM and attended label features from GAT.  
Event sequence processing also tries to derive dependencies between different event types and consider specific attention mechanisms~\cite{mei2021transformer}.
The closest to our LANET approach~\cite{liu2021gated} explores the relationship between different series for multivariate time series classification. The authors propose using attention in step-wise and channel-wise fashion to produce embeddings, which are then forwarded by a classification head. 

\paragraph{\textbf{Sequential recommendation systems.}}
One more close neighbor of our problem statement is the problem of sequential recommendation system construction~\cite{quadrana2018sequence, wang2019sequential}.
In this case, we have many possible labels, and we should sort them by probability of occurrence next time.
Typically, the estimation of embeddings for all possible labels/items is a part of a pipeline.
Existing approaches use neural networks for sequential data such as LSTM~\cite{wu2017recurrent} as well as attention mechanism~\cite{wang2017perceiving}.
We want to highlight the statement related to the usage of only recent past data for prediction~\cite{li2020time}.
However, millions of possible labels typically lead to more classic techniques in this area with specific loss functions and methods.

\paragraph{\textbf{Next basket recommendation.}}

The next relevant problem is the next basket recommendation. 
This formulation is similar to ours, so we also considered many approaches and ideas when analyzing our research area. The authors in~\cite{ariannezhad2023personalized} proposed a personalized model that captures short-term dependencies within a temporary set of products, as well as a long-term one based on historical user information. Also, in~\cite{yannam2023hybrid}, to connect local and global user information, a hybrid method based on an autoencoder for context extraction and RNN for understanding the dynamics of changing interests is proposed. To overcome similar problems, a graph-based hyperedge-based attention network~\cite{song2023hgat} is being created for the following recommendation.
In this formulation of the problem, there is difficulty working with a dictionary of product categories since they number thousands of values; ~\cite{van2023next} uses GRU to predict the next basket, which is easily scaled to a large assortment. 



\section{Methodology} 
\label{sec:methodology}

This section presents the formalization of the temporal sets prediction problem and the description of our LANET method that addresses it effectively. The overall architecture of LANET is presented in Figure~\ref{fig:transf}. We expand in LANET parts consolidation of historical information into joint label representations, application of Transformer encoder, and obtaining a vector of confidence scores. 

\subsection{Temporal Sets Prediction}

In event sequence theory, each event is characterized by one categorical label and a timestamp. In practice, there are lots of available event sequences related to different users with their underlying development patterns. When dealing with such data structure, the wide-spread goal is to capture user- and general-level hidden sequence regularities to predict future behavior. Mostly, an event is attributed not with a single label but with some set of labels. It is a more general and realistic problem statement to consider the possibility of a time moment being concurrently associated with various marks. For instance, engaging a number of services in the app, purchasing several items in the online store, or conducting various kinds of transactions over some period of time. Therefore, the transition from temporal events to temporal sets can be viewed as an act of generalization. In what follows, we treat \textbf{Temporal Sets} as a sequence of event-related timestamped sets composed of an arbitrary number of labels. In turn, \textbf{Temporal Sets Prediction} is a problem of predicting a label set tied to the next event on the basis of an observed sequence of event-associated sets.  

The problem of temporal sets prediction can be formalized as follows. Let $\mathcal{U}=\{u_1, u_2, \dots, u_N \}$ be the collection of $N$ users. Each user $i, 1 \leq i \leq N$, is bound with a sequence of temporal sets $\mathcal{S}_i = \{s_i^1, s_i^2, \dots, s_i^T\}$, where $T$ is a number of the observed timestamps. A set $s_i^j, 1 \leq i \leq N, 1 \leq j \leq T$, is a collection of an arbitrary number of labels sampled from a vocabulary $\mathcal{Y}=\{y_1, y_2, \dots, y_M\}$ of size $M$. Given a sequence of historical sets $\mathcal{S}_i = \{s_i^1, s_i^2, \dots, s_i^T\}$ for user $u_i \in \mathcal{U}$, where each set $s_i^j \subset \mathcal{Y}$, the goal of temporal sets prediction problem is to predict the subsequent label set $\hat{s}_i^{T+1}$, that is,

$$\hat{s}_i^{T+1} = g(s_i^1, s_i^2, \dots, s_i^T, \mathbf{W}), $$ 

where $\mathbf{W}$ relates to trainable parameters of function $g$. Function $g$ should be able to grasp the consecutive development of sets in a sequence $\mathcal{S}_i$ as well as label interaction within each set $s_i^j$.

\subsection{Our LANET approach}
The principal aspects of the temporal sets prediction problem are the time-evolving nature of set series and the complex inner organization of individual sets. Notably, these peculiarities are interconnected and complementary, requiring a joint record. Mindful of the importance of their concurrent treatment, we propose a model LANET that is targeted at such a challenge. In particular, we propose to calculate self-attention between specifically designed representations of historical information. Such representations encompass the knowledge of the time of the events happening and the label composition of each event-related set. The usage of the self-attention mechanism over constructed representations enables the identification of time- and label-aware relationships. Finally, we apply affine transformations to the updated representations at the output of self-attention to get a vector of confidence scores for the next event labels. 

\paragraph{Representation of historical information in LANET.}
First of all, we want to effectively aggregate past information on event times and set structures for the sequence $\mathcal{S}_i = \{s_i^1, s_i^2, \dots, s_i^T\}$. Let $\mathbf{X} \in \mathbb{R}^{M \times D}$ denote the embedding matrix of all labels from the vocabulary $\mathcal{Y}=\{y_1, y_2, \dots, y_M\}$, where $D$ is a dimension of embedding vectors. The parameters of the matrix $\mathbf{X}$ are initialized from the standard normal distribution and later updated in the training process. An important step is the construction of time representations. Each set $s_i^j$ is connected with time $t_j$. The countdown of time starts from one common point for all users. For each timestamp $j, 1 \leq j \leq T$, we establish temporal embedding $\mathbf{t}_j \in \mathbb{R}^{D}$, as it is done in \cite{shou2023concurrent}:

$$\mathbf{t}_j^{(d)} = \begin{cases}
                       \cos{(t_j / 10000^{\frac{d-1}{D}})}, & \text{if} \; d \; \text{is odd},\\
                       \sin{(t_j / 10000^{\frac{d}{D}})}, & \text{if} \; d \; \text{is even},
\end{cases} $$
where $d, 1 \leq d \leq D$, is a component of a vector of dimension $D$. After defining representation for each time moment $t_j, 1 \leq j \leq T$, we aggregate all time-related knowledge from sequence $\mathcal{S}_i = \{s_i^1, s_i^2, \dots, s_i^T\}$ into matrix $\textbf{Z} \in \mathbb{R}^{M \times D}$. The $m$-th row, $1 \leq m \leq M$, of matrix $\textbf{Z}$, denoted as $\textbf{Z}^{(m, :)}$, is equal to the sum of embeddings of timestamps, in which label $y_m \in \mathcal{Y}$ appears as a member of set:

$$\textbf{Z}^{(m, :)} = \sum_{j | y_m \in s_i^j} \mathbf{t}_j$$

If label $y_m$ is not encountered in any set of the sequence $\mathcal{S}_i = \{s_i^1, s_i^2, \dots, s_i^T\}$, then the $m$-th row of matrix $\textbf{Z}$ will consist of all zeros. Hence, in the case of meeting label $y_m$ in several sets of the sequence $\mathcal{S}_i$, the corresponding $m$-th row of matrix $\textbf{Z}$ will be the sum of all relevant time embeddings for this particular label.

The united representation of sequence $\mathcal{S}_i = \{s_i^1, s_i^2, \dots, s_i^T\}$ is a concatenation of defined matrices, embodying time and set structure information:

$$ \mathbf{G} = \mathbf{X} \oplus \mathbf{Z} $$

The rows of resulting matrix $\mathbf{G} \in \mathbb{R}^{M \times 2D}$ are regarded as joint representations of corresponding labels. Namely, $m$-th row of matrix $\mathbf{G}$ is a joint view of label $y_m$. The designed representation of each label includes its view, expressed in $\mathbf{X}$, and part responsible for time-aware interaction with other labels, found in $\mathbf{Z}$.

\paragraph{Learning relations via self-attention in LANET encoder.}
We define the joint label representations as rows of matrix $\mathbf{G}$, which involve self-oriented label information as well as time-aware knowledge of label interrelationships. For the encouragement of further relation capturing, we apply the self-attention mechanism over the matrix $\mathbf{G}$ to get its updated version $\mathbf{\tilde{G}}$:

$$\mathbf{\tilde{G}}=\text{softmax}(\frac{\mathbf{Q}\mathbf{K^T}}{\sqrt{2D}})\mathbf{V}, $$

where $\mathbf{Q}, \mathbf{K}, \mathbf{V}$ are query, key, and value matrices, which are linear transformations of matrix $\mathbf{G}$. The main block of LANET architecture consists of several transformer encoder layers with multi-head self-attention. Leveraging self-attention, we fuse historical records expressed through joint-label views and emphasize essential interactions. The updated label representations are infused with retrospective time- and set structure-aware information. 

\paragraph{LANET prediction layer.}
Finally, the updated representations $\mathbf{\tilde{G}} \in \mathbb{R}^{M \times 2D}$ take part in obtaining the confidence scores for all labels to be included in the next-event set:

$$\mathbf{\hat{f}} = \text{sigmoid}(\mathbf{\tilde{G}} \mathbf{W}^{\text{out}} + b^{\text{out}}), $$

where $\mathbf{\hat{f}} \in \mathbb{R}^{M}$ is a confidence score vector of size of label vocabulary, $\mathbf{W}^{\text{out}} \in \mathbb{R}^{2D \times 1}$ and $b^{\text{out}} \in \mathbb{R}$ are trainable parameters of the prediction layer. We use the sigmoid activation function to make confidence scores lie in the $[0, 1]$ range. Therefore, the $m$-th component, $1 \leq m \leq M$, of confidence vector $\mathbf{\hat{f}} \in \mathbb{R}^{M}$ is associated with a prospect of label $y_m$ to become a part of the predicted set $\hat{s}_i^{T+1}$.

\paragraph{LANET learning process.}
The output of the LANET prediction layer is a confidence score vector $\mathbf{\hat{f}} \in \mathbb{R}^{M}$. Vector $\mathbf{\hat{f}}$ provides the basis for predicting the composition on the next-event set $\hat{s}_i^{T+1}$. For model training and validation, we use a real next set $s_i^{T+1}$ as a ground truth. LANET is trained in end-to-end fashion, taking the historical sequence 
 $\mathcal{S}_i = \{s_i^1, s_i^2, \dots, s_i^T\}$ as an input and producing a vector of confidence score $\mathbf{\hat{f}}$ as an output. We adopt the following loss function:
 
 $$
\mathcal{L}_i = -\frac{1}{M}\sum_{m=1}^M \left(\mathbf{I}_m \log{\mathbf{\hat{f}}}^{(m)} + 
 \mathbf{I}_m^{'} \log{(1-\mathbf{\hat{f}}}^{(m)}) \right), 
$$

where $\mathbf{I}_m  = \mathbf{I}\{ y_m \in s_i^{T+1} \}$ is an indicator function of label $y_m$ to be a member of a set $s_i^{T+1}$, while $\mathbf{I}_m^{'}$ is an indicator function with the opposite condition $\mathbf{I}_m^{'}  = \mathbf{I}\{ y_m \notin s_i^{T+1} \}$. We denote the $m$-th component of the predicted confidence score vector $\mathbf{\hat{f}}$ as $\mathbf{\hat{f}}^{(m)}$. The formula of loss function $\mathcal{L}_i$ is given for the case when we consider only one user $u_i$. In view of all available users $\mathcal{U}=\{u_1, u_2, \dots, u_N\}$, we minimize the sum of all user-related loss components $\mathcal{L} = \sum_{i=1}^N \mathcal{L}_i$ in the training process.   

In the training dataset for LANET, the bundle of the set $s_i^{T+1}$ as a ground truth and a sequence $\{s_i^1, s_i^2, \dots, s_i^T\}$ as input is not the only one training example that is drawn from the user sequence $\mathcal{S}_i$. To increase the amount of training data, we also leverage all intermediate sets in a sequence as a ground truth and preceding sets as the model's input. Thus, $s_i^j, 2 \leq j \leq T+1, $ are taken as target sets and the subsequences $\{s_i^1, \dots, s_i^{(j-1)}\}$ as corresponding inputs.

\section{Experiments} 
\label{sec:experiments}

\begin{table*}[h!] 
\centering
\small
\caption{Comparison of our LANET approach with the existing models for temporal sets prediction on four datasets. Best values are highlighted, and second-best values are underlined.
}
\vspace{2pt}
\begin{tabular}{cccccccc}
    \hline
    Dataset & Model &Micro F1$\uparrow$& Macro F1$\uparrow$ & Weighted F1$\uparrow$ & Weighted ROC-AUC$\uparrow$ & Hamming Loss$\downarrow$\\ \hline
    \multirow{5}{*}{Synthea}
    & SFCNTSP &0.2369	$\pm$ 0.0156& 0.0587 $\pm$ 0.0069 & 0.1656 $\pm$ 0.0194 & 0.6655 $\pm$ 0.0077 & 0.0212 $\pm$ 0.0005\\
    & DNNTSP &0.3893 $\pm$ 0.0181& 0.1288 $\pm$ 0.0058 & 0.2982 $\pm$ 0.0132 & 0.7070 $\pm$ 0.0076 & 0.0183 $\pm$ 0.0006 \\
    & GPTopFreq &0.4100	$\pm$ 0.0042& 0.1312 $\pm$ 0.0097 & 0.3286 $\pm$ 0.0083 & 0.7229 $\pm$ 0.0093 & 0.0183 $\pm$ 0.0003 \\
    & TCMBN &\underline{0.4551	$\pm$0.0126} & \underline{0.1522 $\pm$ 0.0023} & \underline{0.3538 $\pm$ 0.0080} & \underline{0.8347 $\pm$ 0.0047} & \textbf{0.0173 $\pm$ 0.0004} \\ 
     & LANET(ours) &\textbf{0.5277 $\pm$ 0.0098}& \textbf{0.2724 $\pm$ 0.0122} & \textbf{0.4704 $\pm$ 0.0071}  & \textbf{0.9026 $\pm$ 0.0018} & \underline{0.0175 $\pm$ 0.0005} \\\hline
    \multirow{5}{*}{Mimic III} 
    & SFCNTSP &0.4298 $\pm$0.0032& 0.2338 $\pm$ 0.0071 & 0.3791 $\pm$ 0.0081 & 0.7034 $\pm$ 0.0024 & 0.0377 $\pm$ 0.0004 \\
    & DNNTSP &0.4362 $\pm$0.0025& 0.2552 $\pm$ 0.0034 & 0.3928 $\pm$ 0.0030 & 0.6926 $\pm$ 0.0003 & 0.0365 $\pm$ 0.0003\\ 
    & GPTopFreq &0.4405	$\pm$0.0070& \underline{0.3089 $\pm$ 0.0039} & 0.4291 $\pm$ 0.0073 & 0.6912 $\pm$ 0.0028 & 0.0398 $\pm$ 0.0005 \\
    & TCMBN &\underline{0.5419	$\pm$0.0151}& 0.2603 $\pm$ 0.0276 & \underline{0.4979 $\pm$ 0.0180} & \underline{0.8670 $\pm$ 0.0095} & \underline{0.0305 $\pm$ 0.0008} \\ 
    & LANET(ours) &\textbf{0.8218$\pm$	0.0211} & \textbf{0.7408 $\pm$ 0.0377} & \textbf{0.8214 $\pm$ 0.0224}   & \textbf{0.9852 $\pm$ 0.0023} & \textbf{0.0220 $\pm$ 0.0001} \\\hline
    \multirow{5}{*}{DC}
    & SFCNTSP &0.1081 $\pm$0.0058& 0.0831 $\pm$ 0.0047 & 0.0886 $\pm$ 0.0054 & 0.7014 $\pm$ 0.0024 & 0.0077 $\pm$ 0.0001 \\
    & DNNTSP &0.0356 $\pm$ 0.0041& 0.0254 $\pm$ 0.0031 & 0.0259 $\pm$ 0.0027 & 0.6784 $\pm$ 0.0000 & \underline{0.0074 $\pm$ 0.0000}\\ 
    & GPTopFreq &0.1623	$\pm$ 0.0019& 0.1449 $\pm$ 0.0027 & 0.1525 $\pm$ 0.0019 & 0.6533 $\pm$ 0.0022 & 0.0083 $\pm$ 0.0001 \\
    & TCMBN & \underline{0.2288 $\pm$ 0.0153}& \underline{0.1788 $\pm$ 0.0136} & \underline{0.1968 $\pm$ 0.0134} & \underline{0.8932 $\pm$ 0.0048} & \textbf{0.0073 $\pm$ 0.0001} \\ 
    & LANET(ours) &\textbf{0.5608 $\pm$ 0.0097}& \textbf{0.5473 $\pm$ 0.0134} & \textbf{0.5498 $\pm$ 0.0137} & \textbf{0.9941 $\pm$ 0.0004} & 0.0085 $\pm$ 0.0002 \\ \hline
    \multirow{5}{*}{Instacart}
    & SFCNTSP &0.2756 $\pm$ 0.0140& 0.0283 $\pm$ 0.0031 & 0.1672 $\pm$ 0.0112 & 0.6852 $\pm$ 0.0448 & 0.0581 $\pm$ 0.0004 \\
    & DNNTSP &\underline{0.4476 $\pm$0.0021}& \underline{0.2623 $\pm$ 0.0041} & \underline{0.4160 $\pm$ 0.0009} & 0.7913 $\pm$ 0.0004 & 0.0541 $\pm$ 0.0002 \\ 
    & GPTopFreq &0.4376	$\pm$0.0061& 0.2581 $\pm$ 0.0035 & 0.4087 $\pm$ 0.0079 & 0.7736 $\pm$ 0.0039 & \underline{0.0529 $\pm$ 0.0008}  \\
    & TCMBN &0.4192	$\pm$0.0064& 0.1577 $\pm$ 0.0066 & 0.3687 $\pm$ 0.0065 & \underline{0.8187 $\pm$ 0.0030} & 0.0530 $\pm$ 0.0005 \\ 
    & LANET(ours) &\textbf{0.6253 $\pm$ 0.0026} & \textbf{0.4916 $\pm$ 0.0082} & \textbf{0.6159 $\pm$ 0.0029}  & \textbf{0.9445 $\pm$ 0.0008} & \textbf{0.0474 $\pm$ 0.0003}\\ \hline
    \end{tabular}
\centering
\label{table:diff_data}
\end{table*}

In this section, we present the performance comparison of our LANET approach with the existing models for temporal sets prediction problem. Besides, we perform a thorough ablation study that reveals insights into LANET working details. The code for LANET is available at GitHub repository\footnote{\url{https://github.com/adenshulga/LANET}}.

\subsection{Datasets}
After analysis of the works devoted to models for temporal sets prediction, we identify four frequently used datasets:
\begin{itemize}
\item \textbf{Dunnhumby-Carbo (DC)}~\cite{Dunnhumby}: This dataset includes transactional data of households in a retail store over two years. Here, sets are products assigned to one transaction.
    
\item \textbf{Mimic III}~\cite{mimic3}: It consists of the medical records for patients from intensive care. The patient-related event constitutes a hospital admission time and a set of disease classification codes.
    
\item \textbf{Instacar}t~\cite{instacart}: The Instacart dataset comprises records of users' product orders. Each event is described by a time of purchase and a set of product labels.
    
\item \textbf{Synthea}~\cite{synthea}: This is synthetically generated EHR data with simulated medical events, similar to the MIMIC III dataset.
\end{itemize}

Statistics of these datasets are given in Table~\ref{tab:multi_datasets}. We provide the overall number of sets in each dataset (\#Sets), the median set size (MdnSS), the maximum set size (MaxSS), the size of label vocabulary (Vocab), the mean length of historical sequences (MnLen), and the number of available sequences (\#Seqs).

\begin{table}[ht!] 
\caption{Statistics of the datasets for temporal sets prediction.}
\vspace{2pt}
\begin{tabular}{p{1.2cm}p{0.9cm}p{0.8cm}p{0.8cm}p{0.55cm}p{0.8cm}p{0.73cm}}
\hline
Dataset & \#Sets & MdnSS & MaxSS & Vocab & MnLen & \#Seqs  \\
\hline 
Synthea & 108 439 & 2 & 13 & 232 & 44.1 & 2459 \\
Mimic III & 17 849 & 5 & 23  & 169 & 2.7 & 6636 \\
Synthea & 108 439 & 2 & 13 & 232 & 44.1 & 2459 \\
DC & 121 165 & 1 & 9  & 217 & 3.6 & 33895 \\
Instacart & 115 604 & 6 & 43  & 134 & 16.5 & 7000 \\
\hline
\end{tabular}
\label{tab:multi_datasets}
\end{table}

\begin{figure*}
\centering
\begin{minipage}{.3\textwidth}
\centering
\includegraphics[width=\linewidth]{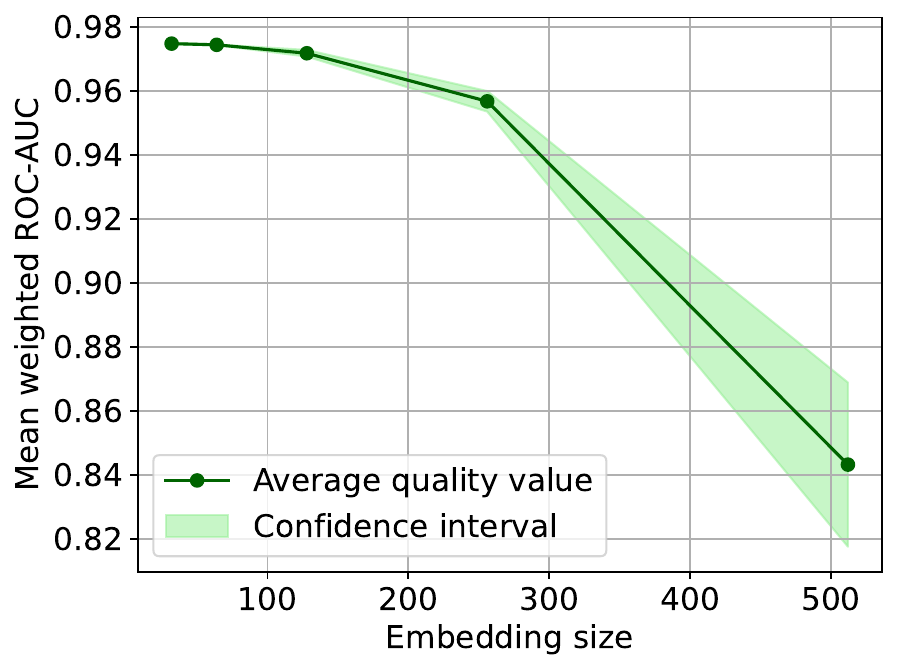}
\caption{The dependence of LANET quality on the embedding size.}
\vspace{10pt}
\label{fig:test1_emb}
\end{minipage}\hfill
\begin{minipage}{.3\textwidth}
\centering
\includegraphics[width=\linewidth]{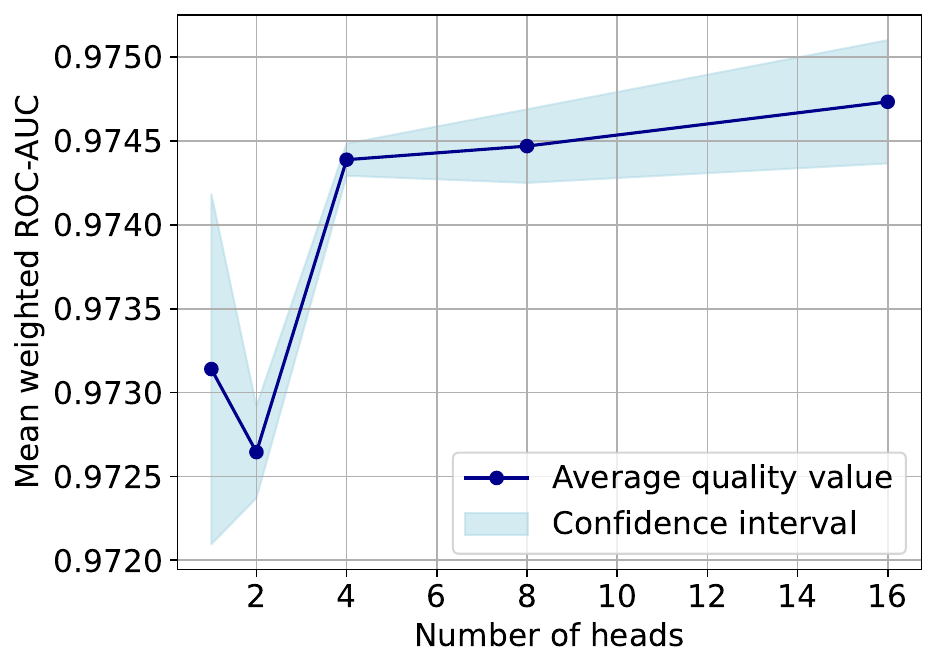}
\caption{The dependence of LANET quality on the number of heads.}
\vspace{10pt}
\label{fig:test2_head}
\end{minipage}\hfill
\begin{minipage}{.3\textwidth}
\centering
\includegraphics[width=\linewidth]{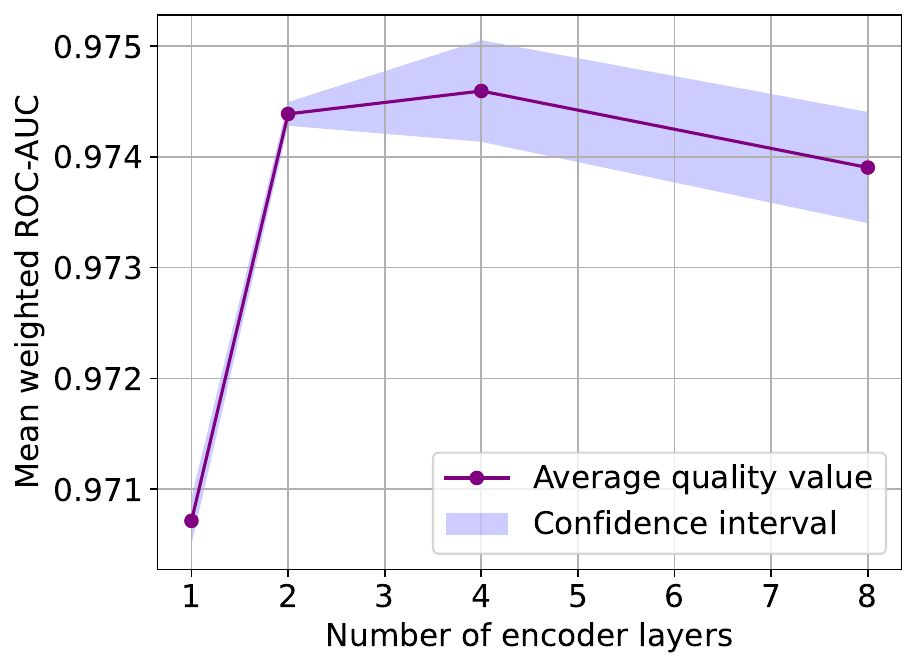}
\caption{The dependence of LANET quality on the number of encoder layers.}
\vspace{10pt}
\label{fig:test3_layer}
\end{minipage}
\end{figure*}

\subsection{Compared Methods}
The following methods are compared with our LANET approach:

\begin{itemize}
\item \textbf{GPTopFreq} is a frequency-based baseline, inspired by~\cite{li2023next}. This method evaluates the frequencies of each label occurrence in the whole dataset and in the user-related history. Then, for each label, GPTopFreq takes the maximum of ``general” and ``personal" frequencies and uses it as the predicted probability.

\item \textbf{DNNTSP\footnote{\url{https://github.com/yule-BUAA/DNNTSP}}} model is described in~\cite{zhang2020multi}. DNNTSP constructs a co-occurrence frequency graph, performs weighted graph convolutions on it to learn element relationships, utilizes an attention-based module to learn the temporal dependency of elements in sets, and uses a gating mechanism to fuse static and dynamic information about elements. 
 
\item \textbf{SFCNTSP\footnote{\url{ https://github.com/yule-BUAA/SFCNTSP}}} is a model for temporal sets prediction presented in~\cite{yu2023predicting}. It comprises four consequent modules, namely Simplified Fully-Connected Networks, that learn inter and intra-set dependencies, intra-embedding channel correlations, and user representations.

\item \textbf{TCMBN\footnote{\url{https://github.com/xshou1990/TCMBN}}} model idea is given in~\cite{shou2023concurrent}. TCMBN leverages Transformer-based architecture to capture probabilistic dependency between elements in sets via neural density estimation of parameters of Bernoulli mixture and temporal dependency between sets via attention.
\end{itemize}

We take these models because they are pretty recent in temporal sets prediction and demonstrate high performance. Their hyperparameters for different datasets are set to the values specified by the authors. 

\subsection{Implementation details}
Our LANET model consists of several transformer encoder layers with multi-head self-attention. The number of layers is equal to $2$ in all cases, while the number of self-attention heads ranges from $4$ to $6$, depending on the particular dataset. As a basis, we take the Transformer layer implementation from PyTorch~\cite{transformer_encoder_torch}. We apply dropout with the probability of $0.2$ directly to the output of the transformer encoder block. LANET quality dependence on the model hyperparameters will be presented in Section~\ref{sec:qual}. For the training procedure, we use the Adam optimizer with an initial learning rate of $0.001$. For the scheduler, we adopt ``reduce on Plateau” strategy with patience of $10$ epochs and factor of $0.9$.

\subsection{Validation metrics}
The original datasets are divided into train, validation, and test sets. Splits are performed on users. Thus, time periods in the train, valid, and test parts overlap.
We take 60\% of the data samples for model training, 20\% for validation, and 20\% for testing. All experiments are launched with five different random seeds; the mean and standard deviation of the results are calculated.

Evaluation of temporal sets prediction is similar to validation of multi-label classification, so we use well-established and comprehensive metrics from multi-label domain~\cite{wu2017unified} and metrics from the relevant works~\cite{shou2023concurrent} of the considered area of temporal sets. Thus, we employ Hamming Loss, Weighted ROC-AUC, Weighted F1, Micro F1, and Macro F1 metrics for ultimate quality assessment. Meantime, the calculation of micro-F1, macro-F1, and Weighted F1 implies operation with the predicted label sets, not with the label confidence scores. In this regard, the transition from output scores to the predicted label sets is done by comparison of the label-related confidence scores with certain thresholds. These thresholds are calculated on the validation set by optimization of F1-score for each label separately.


\subsection{Main results} 
\label{sec:comp_qual}

Metrics for comparison of our LANET approach with the established models for temporal sets prediction problem are presented in Table~\ref{table:diff_data}. LANET demonstrates top-$1$ performance on all datasets, substantially surpassing its competitors. A huge performance gap is observed on the DC, which can be connected with a vast number of available sequences for training in this dataset or with the specific set structures. The closest competitor for LANET is the TCMBN model, which is also based on transformer architecture. The results indicate that the crucial moment is the treatment of the historical information at the model entrance rather than its processing afterwards. LANET successfully copes with this challenge and shows an absolutely different level of performance. Interestingly, the statistical baseline GPTopFreq demonstrates a higher quality than the deep neural network models in some cases. Such phenomenon is also mentioned in~\cite{li2023next}.

\subsection{Ablation study}
\label{sec:qual}

We investigate the dependence of LANET performance on its major hyperparameters. Unless otherwise specified, MIMIC III dataset is under consideration.

\paragraph{\textbf{Contribution of time information.}}
Each set in a sequence is attributed with a timestamp, which takes part in obtaining time representations. We decide to contemplate the contribution of the time component to model performance. So, we omit temporal information from LANET by substituting time representations with the constant vector. Such a vector indicates the particular label's presence in the user history, neglecting all time dependencies. The metric drops as a result of this modification are given in Table~\ref{table:attents}. However, even after the exclusion of the time-aware views from LANET, it still demonstrates elevated performance due to the efficient processing of similar frequency-based history representation.


\begin{table}[ht!]
\centering
\small
\caption{The contribution of temporal information into LANET performance. Metric drops in case of time omission are provided for Weighted F1 and Weighted ROC-AUC.}
\vspace{2pt}
\begin{tabular}{p{1.3cm}p{1.3cm}p{2.2cm}p{2.2cm}}
    \hline
    Dataset & Model & F1 & ROC-AUC \\
    \hline
    \multirow{2}{*}{ Synthea } 
    & No time & 0.3890 $\pm$ 0.0162 & 0.8810 $\pm$ 0.0023 \\
    & LANET & 0.4704 $\pm$ 0.0071  & 0.9026 $\pm$ 0.0018 \\
     \hline
    \multirow{2}{*}{Mimic III}  
    & No time & 0.7644 $\pm$ 0.0023 & 0.9775 $\pm$ 0.0001 \\
    & LANET & 0.8214 $\pm$ 0.0224 & 0.9852 $\pm$ 0.0023 \\
     \hline
    \multirow{2}{*}{ DC }  
    & No time & 0.4316 $\pm$ 0.0044 & 0.9906 $\pm$ 0.0000 \\
    & LANET & 0.5498 $\pm$ 0.0137 & 0.9941 $\pm$ 0.0004 \\
     \hline
    \multirow{2}{*}{ Instacart }  
    & No time & 0.5277 $\pm$ 0.0032 & 0.9145 $\pm$ 0.0004 \\
    & LANET & 0.6159 $\pm$ 0.0029 & 0.9445 $\pm$ 0.0008 \\
     \hline
    \end{tabular}
\label{table:attents}
\end{table}

\paragraph{Dependence of LANET performance on embedding size.}

An essential part of our model is the utilization of learnable embeddings for managing temporal sets. For this reason, it is necessary to examine the influence of embedding dimensionality on LANET metrics because this parameter is directly related to a model capacity. The dimension of joint representations before transferring into the transformer encoder bock equals $2D$. The effect of changing the values of $D$ is presented in Figure~\ref{fig:test1_emb}. From it, we can conclude that LANET struggles to learn representations of high dimensions effectively.

\paragraph{Dependence of LANET performance on number of heads in attention layers.}
The usage of several heads in the attention layers allows the model to account for multiple distinct dependencies, dedicating an individual head to grasping the specific pattern. Figure~\ref{fig:test2_head} confirms that the more significant number of adopted heads leads to quality enhancements. However, resource consumption grows alongside the increase in the head quantity.
 
\paragraph{Dependence of LANET performance on a number of encoder layers.}
The hyperparameter of the number of encoder layers is responsible for the capability to recognize complex relationships within data. Figure~\ref{fig:test3_layer} demonstrates that there exists an optimal number of layers for solving the considered problem. The further increase in the number of layers brings in the failure to train the effective model.  

\paragraph{\textbf{Graph interpretation of attention weights.}}

\begin{figure*}
\centering
\begin{minipage}{.22\textwidth}
\centering
\includegraphics[width=\linewidth]{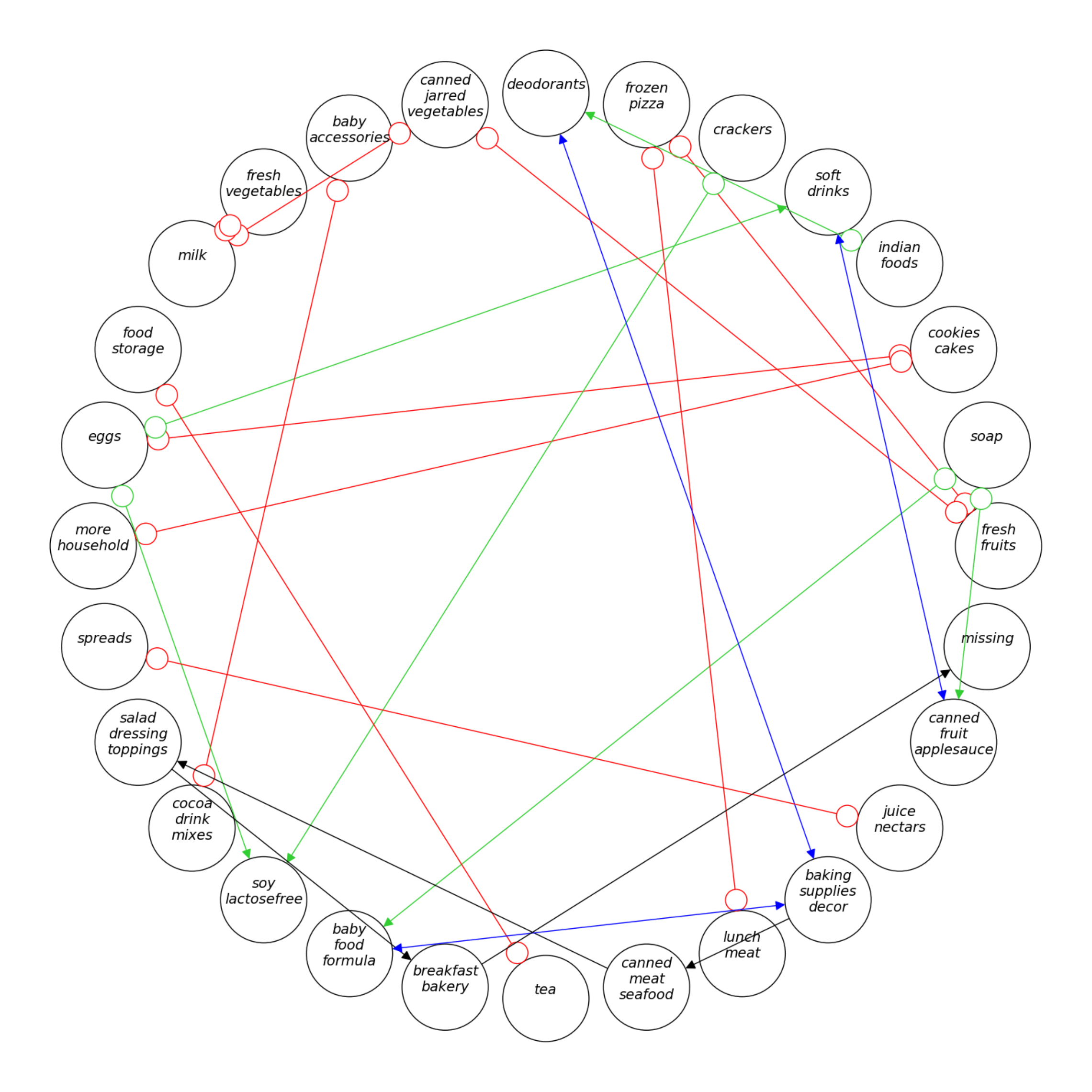} 
\label{fig:test1}
\end{minipage}
\begin{minipage}{.22\textwidth}
\centering
\includegraphics[width=\linewidth]{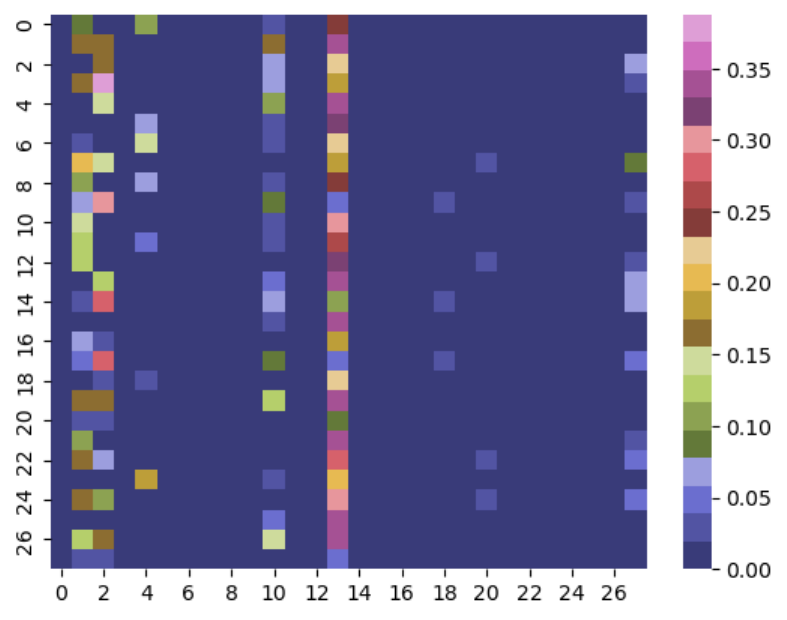} 
\label{fig:test2}
\end{minipage}
\begin{minipage}{.22\textwidth}
\centering
\includegraphics[width=\linewidth]{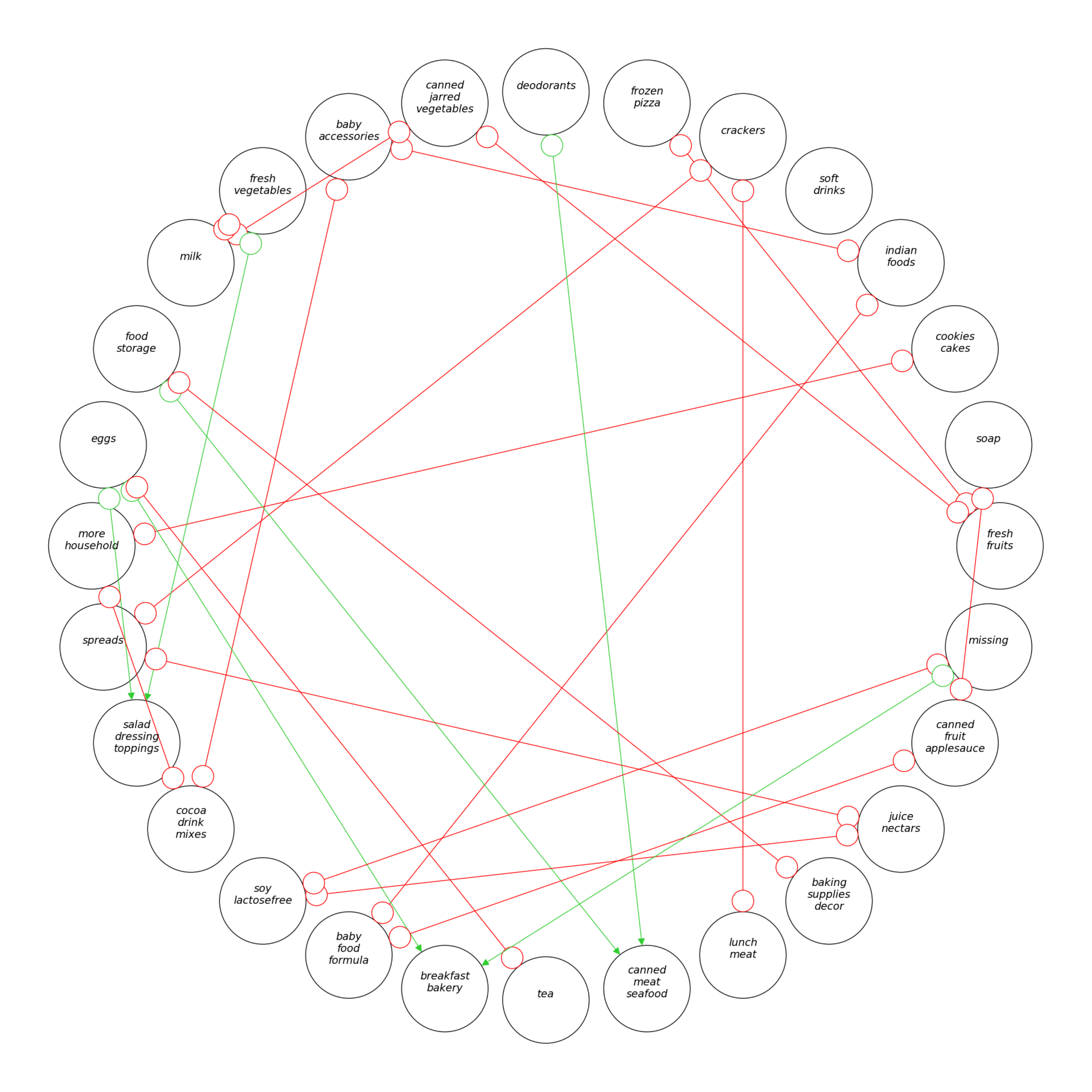} 
\label{fig:test3}
\end{minipage}
\begin{minipage}{.22\textwidth}
\centering
\includegraphics[width=\linewidth]{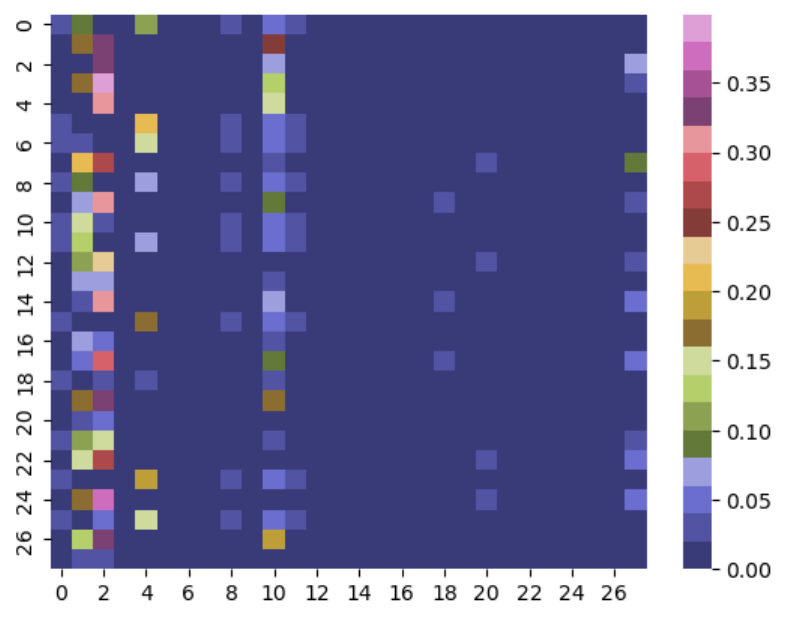} 
\label{fig:test4}
\end{minipage}\caption{Interpreting the relationship of labels using the attention layer. On the left is a picture showing the relationship between a subset of labels and their verbal interpretation. Next to the graph is a heatmap, which illustrates the relationship of all possible labels of the Instacart dataset. On the right are the modified graphs, which are obtained as a result of removing the label with the highest attention weight in all possible values and the corresponding distribution of weights in the heatmap. The data is obtained from the dataset Instacart.}
\vspace{10pt}
\label{fig:ablation2}
\end{figure*}
An essential part of the resulting architecture is the encoder layer, which includes the attention layer. Attention, in turn, indicates the degree of relevance of the relationship between the labels, which is significant for further model prediction. We select the most relevant labels for a selection in Instacart to identify the causal explanations of label predictions. The Figure~\ref{fig:ablation2} on the left shows the heatmap for their relationships. We notice that the attention matrix clearly dominates of the labels encountered in the sequence over those that are not in it, which is clearly expressed through the weights. Looking deeper, we see that small-scale variations in attention describe the connection between particular event types. 

Furthermore, we consider the most relevant labels for a sampling. The figure on the left shows the heatmap for their relationships. To generate causal explanations, we needed a graph visualization of the attention scales for individual labels. This is an idea behind framework CLEANN~\cite{rohekar2024causal}, which proposes a method to extract causal relationships as a partial ancestral graph (PAG)~\cite{richardson2002ancestral}. So, to form the graph, we looked at one of the users and the corresponding historical information about the labels. Using the pretrained LANET model, we obtained the attention weights fed into the CLEANN algorithm. 

The left visualization of the graph in Figure~\ref{fig:ablation2}  has several types of connections:
\begin{itemize} 
    \item The red lines indicate the proximity of the labels inside the graph;
    \item The blue connections are more complex, this is a bidirectional interaction between the labels in the graph;
    \item Black means that the label is the parent for the subsequent;
    \item The greens, on the contrary, are children.
\end{itemize}
In the first case, complex and intricate relationships between labels have developed. For example, if ``canned meat seafood" is the parent, you will generate ``salad dressing toppings." Some connections may seem counterintuitive to us, but this story is individual for each user when buying goods in the store. 

Moreover, in order to find out and identify the connections, we decided to remove the label with the highest total weight in the attention matrix and look at the redistribution of weights in this case (Figure~\ref{fig:ablation2} on the right). The model turned the attention to a variety of other labels. The PAG demonstrates a changed picture, where all the blue and black edges of the graph have disappeared, which corresponds to a more complex and oriented connection than a simple  ``neighborly" one. The correlation between labels has become lower. Moreover, ``canned meat seafood" changed its behavior. It has become a subsidiary and no longer has connections with anyone, which affects the predictive ability of this label for the next time step. This exploration indicates that the best predictive capabilities of LANET mainly depend on the model's ability to detect relationships between labels rather than on building a work with time and the order in which baskets are placed.

\section{Conclusion}

In this work, we consider the problem of temporal sets prediction: given the history of timestamped sets comprised of an arbitrary number of categorical labels, the goal is to predict the collection of labels for the next event. To solve this problem, we propose the LANET model. LANET is remarkable for its early effective aggregation of historical information into vector representations, not encountered in other existing models. The specific view on the available information enables further effective capturing of time and label interdependencies. Our method demonstrates the best performance on four reviewed datasets, surpassing SOTA approach and providing improvement of $65\%$ in terms of Weighted F1 on one of the datasets. As for the limitations, LANET shows consistently strong results, specifically on datasets with label vocabulary sizes of $100-200$. The adaptation to the recommendation setting, in which item vocabulary size can reach thousands, or to the setup with the much smaller vocabulary of $5-20$ are open questions. Besides, the issue that is worthy of consideration is the study of effects from the reduction of event sequences to temporal sets. Such contraction can be done by choosing an appropriate time period for group formation but may introduce unexpected findings in event sequence tasks. In addition to, the proposed approach naturally fits into the paradigm of self-supervised learning and can serve as a source of valuable representations for the downstream tasks. 





\section{Acknowledgments}
The research was supported by the Russian Science Foundation grant 20-7110135. The authors would like to thank Evgenia Romanenkova for her careful paper review and valuable advice.

\bibliography{mybibfile}

\end{document}